\documentclass[10pt,twocolumn,letterpaper]{article}

\usepackage{medddc_preprint}
\def\ARXIVPREPRINT{1}

\usepackage[hyphens]{url}
\usepackage{graphicx}
\usepackage{natbib}
\usepackage{caption}
\usepackage{booktabs}
\usepackage{colortbl}
\usepackage{multirow}
\usepackage{amsmath}
\usepackage{amssymb}
\usepackage{tikz}
\usetikzlibrary{arrows.meta,positioning,calc,fit,backgrounds}
\title{MedDDC-Eval: Diagnosis-Decoupled Evaluation of Multi-Turn Medical Consultation Agents}
\author{
  Guofeng Zhang$^{1,*,\dagger}$,
  Yizeng Quan$^{1,2,*}$,
  Huaiyi Fang$^{1}$,
  Jianwei Lv$^{1}$,\\
  Jinyao Liu$^{1,3}$,
  Xunxu Duan$^{1,2}$,
  Lening An$^{1}$,
  Yu Ouyang$^{1,\dagger}$,
  Junfeng Wang$^{1}$\\[0.4em]
  \normalsize $^{1}$Baidu, Inc. \quad
  $^{2}$Peking University \quad
  $^{3}$Zhejiang University
}
\date{}

\newcommand{\arxivauthornotes}{%
  \begingroup
  \renewcommand{\thefootnote}{$*$}%
  \footnotetext{These authors contributed equally.}%
  \renewcommand{\thefootnote}{$\dagger$}%
  \footnotetext{\raggedright Corresponding authors: \texttt{zhangguofeng@baidu.com} and \texttt{ouyangyu@baidu.com}}%
  \endgroup
}

\begin{document}
\maketitle
\arxivauthornotes

\begin{abstract}
Evaluating multi-turn medical consultation agents requires judging the diagnostic support provided by the histories they elicit through interaction. Yet coupled evaluation lets each policy both elicit the history and generate the terminal diagnosis, so a diagnosis score confounds the elicited history with the policy's own terminal diagnosis generator. We introduce \textbf{MedDDC-Eval}, a diagnosis-decoupled evaluation testbed over held-out cases derived from medical records and online consultations. It applies the same frozen shared diagnostic reader to every policy-elicited history, holding terminal diagnosis generation fixed across policies and enabling comparison under the shared diagnostic reader. It reports diagnostic support, information-acquisition coverage, and efficiency. LLM-assisted semantic matching followed by deterministic one-to-one assignment makes the diagnosis--trajectory--efficiency (D/T/E) scores auditable. In a fixed-history audit across eight policies, replacing each policy's own generator with the shared diagnostic reader shifts diagnosis F1 by 2.2--19.0 points and reverses 18\% and 36\% of pairwise orderings on the Record and Dialogue splits. To examine downstream utility, we use standard Group Relative Policy Optimization (GRPO) with a separate training-time reward that targets the same diagnosis and trajectory dimensions. Relative to its Qwen3-32B initialization, the trained policy gains 9.6 and 4.6 aggregate-score points on the held-out Record and Dialogue splits, respectively, and ablating either feedback signal reduces the aggregate score on both. Together, MedDDC-Eval supports comparison under a shared diagnostic reader and evaluation-informed policy development, while complementing end-to-end evaluation when terminal diagnosis generation is also part of the target capability.

\end{abstract}

\section{Introduction}

Multi-turn medical consultation agents do not merely answer from supplied evidence; they help construct the evidence on which diagnosis is based. Unlike static medical QA and diagnosis benchmarks, where a case description provides the evidence in advance~\citep{singhal2023clinicalknowledge,wang2023cmb}, their questions determine which evidence becomes available for diagnosis~\citep{qiao2026medconsultbench,sanghvi2026medxagent,lai2025doctorr1}. They must adapt to patient responses, screen for missing evidence, and decide when a bounded history is sufficient. Evaluation must therefore measure not only the final diagnosis but also the history elicited by the consultation policy.

Recent evaluation suites increasingly make inquiry visible through multimodal and agentic tracks, physician-authored criteria, consultation rubrics, and information-acquisition measures~\citep{ding2025medbenchv4,bedi2025medhelm,tu2025amie,arora2025healthbench,gong2026meddialogrubrics,qiao2026medconsultbench}. Yet a central attribution problem remains. When the evaluated system both elicits the history and generates the terminal diagnosis, its diagnosis score reflects two components: the history elicited by the policy and the interpretation applied by its own terminal diagnosis generator. How can consultation policies be compared without simultaneously comparing their terminal diagnosis generators? The problem is not that terminal generators are necessarily weak; it is that they vary across policies and therefore change the object being compared. A strong generator can compensate for a thin history, while a weaker one can obscure a rich history. Process scores describe parts of the consultation, but the downstream diagnostic value of two elicited histories remains entangled with their terminal diagnosis generators. Coupled scores can therefore misattribute gains, misrank consultation policies, and direct reward design toward a different component than intended.

\begin{figure*}[t]
\centering
\includegraphics[width=0.98\textwidth]{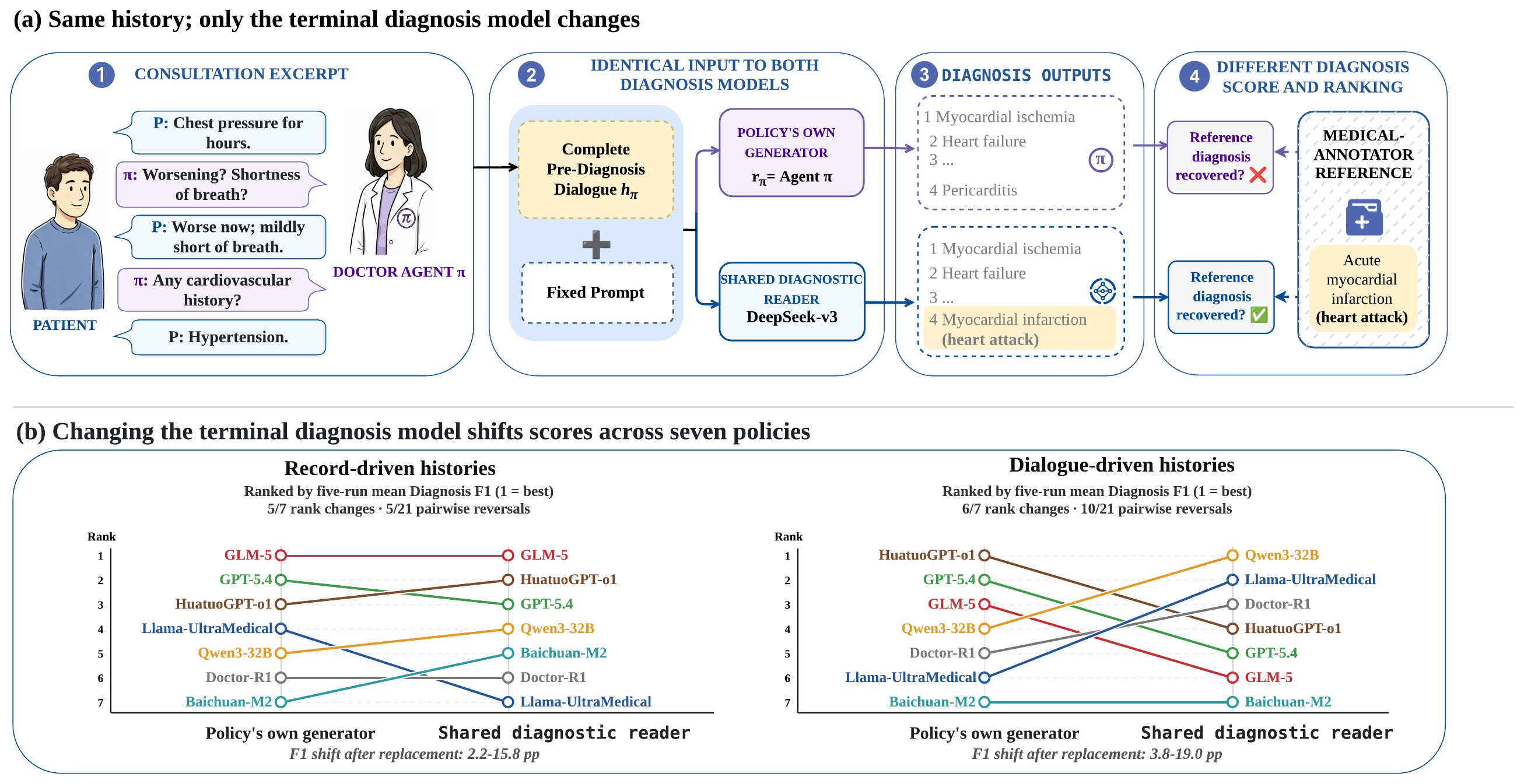}
\caption{Fixed-history terminal-diagnosis intervention. (a) The history, prompt, and decoding settings are fixed; replacing each policy's own terminal diagnosis generator with the shared diagnostic reader changes which reference diagnoses are recovered. (b) The seven-policy view shows 5/21 Record and 10/21 Dialogue pairwise reversals. Exact endpoints and the complete eight-policy audit (5/28 and 10/28) appear in Supplementary Table S7.}
\label{fig:object-vector}
\end{figure*}

Figure~\ref{fig:object-vector} exposes this confound by holding the history, diagnostic prompt, and decoding settings fixed while changing the history-to-diagnosis model. This intervention shifts diagnosis F1 by 2.2--19.0 points and reverses 18\% and 36\% of pairwise policy orderings on the Record and Dialogue splits, respectively, in the complete eight-policy audit. Because the histories are identical, the heterogeneous shifts show that model choice alters between-policy comparisons instead of adding a uniform offset.

Our key idea is simple: hold terminal diagnosis generation fixed across policies and treat the policy-elicited history as the evaluation object. We therefore introduce \textbf{MedDDC-Eval}, a diagnosis-decoupled evaluation testbed over 170 held-out cases derived from medical records and online consultations. A frozen shared DeepSeek-v3 (DS-V3) diagnostic reader~\citep{deepseekai2025deepseekv3} maps every history to a terminal diagnosis, removing each policy's own terminal diagnosis generator as a source of between-agent variation. This control enables comparison of policy-elicited histories under the shared diagnostic reader and complements end-to-end evaluation when terminal generation is also part of the target capability. The testbed reports diagnostic support, information-acquisition coverage against reviewed reference-question targets, and efficiency separately. LLM-assisted semantic matching recognizes free-form paraphrases; deterministic one-to-one assignment prevents duplicate credit and exposes traceable TP/FP/FN counts.

We next ask whether the evaluation dimensions operationalized by MedDDC-Eval can inform model development. In a controlled evaluation-informed study, standard GRPO~\citep{shao2024deepseekmath} post-trains Qwen3-32B~\citep{yang2025qwen3} over interactive rollouts using a separate reward with diagnosis and trajectory feedback plus auxiliary behavioral constraints. Under the frozen held-out evaluator, the trained policy gains 9.6 and 4.6 Total points on the two sources; ablating either feedback signal reduces Total on both.

Our contributions are:
\begin{enumerate}
    \item \textbf{Diagnosis-decoupled evaluation.} MedDDC-Eval compares policy-elicited histories under a frozen shared diagnostic reader; a fixed-history intervention shows that the terminal diagnosis generator changes diagnosis scores and rankings.
    \item \textbf{Auditable elicited-history measurement.} D/T/E characterize diagnostic support, information-acquisition coverage, and efficiency, while semantic matching and one-to-one assignment provide traceable counts.
    \item \textbf{Evaluation-informed policy development.} A controlled Qwen3-32B study uses a separate training-time reward organized around the diagnosis and trajectory dimensions; the trained policy improves on both held-out sources, and removing either feedback signal lowers joint performance.
\end{enumerate}

\section{Related Work}

\subsection{Medical Evaluation and Inquiry}
Conventional static medical QA and diagnosis benchmarks condition on evidence already present in the case~\citep{singhal2023clinicalknowledge,wang2023cmb}. Broader suites such as MedBench v4 and MedHELM include multimodal, agentic, workflow, and clinician-validated tracks~\citep{ding2025medbenchv4,bedi2025medhelm}. Clinical-agent benchmarks instead evaluate sequential decisions, tool use, and workflow execution~\citep{schmidgall2024agentclinic,jiang2025medagentbench,liu2026physicianbench,lee2025fhiragentbench,qiao2026ehrcomplex}. For consultation agents, this distinction changes the evaluation object: the inquiry policy helps construct the diagnostic context on which later decisions depend.

The closest interactive benchmarks make different parts of this process observable. AMIE evaluates a complete conversational inquiry--diagnosis system, while HealthBench and MedDialogRubrics score outputs against physician-authored criteria or consultation rubrics~\citep{tu2025amie,arora2025healthbench,gong2026meddialogrubrics}. MedConsultBench explicitly evaluates information acquisition across the consultation cycle~\citep{qiao2026medconsultbench}; ThReadMed-QA, CP-Bench, and MeDxAgent broaden the setting to patient question threads, challenging patient behavior, and multi-agent consultation~\citep{munnangi2026threadmedqa,li2026cpbbench,sanghvi2026medxagent}. These benchmarks establish inquiry quality as a capability, but an end-to-end diagnosis score still combines the elicited history with the system's terminal generator.

MedDDC-Eval asks a complementary attribution question: how well does a policy-elicited history support diagnosis when every history is interpreted by the same diagnostic reader? DDXPlus also studies evidence acquisition in a controlled differential-diagnosis simulator~\citep{tchango2022ddxplus}; our setting targets open-ended consultation histories and compares them under a frozen shared diagnostic reader. This control holds terminal interpretation fixed across policies, while coupled evaluation remains necessary for judging the complete inquiry--diagnosis pipeline.

\subsection{Multi-Turn RL for LLM Agents}
GRPO is a critic-free grouped-reward method introduced in DeepSeekMath~\citep{shao2024deepseekmath}. Related work studies reproducible and scalable LLM reinforcement learning, as well as finer-grained credit assignment for multi-turn agents~\citep{yu2025dapo,liu2025r1zero,zhang2025agentrl,wei2025turnreward,feng2025gigpo,li2026turnppo,djuhera2026tsr}. Doctor-R1 combines process and terminal rewards to train a medical inquiry agent~\citep{lai2025doctorr1}. Our evaluation-informed training study instead asks whether a separate reward organized around diagnosis and trajectory feedback improves the history elicited by a fixed Qwen3-32B policy family~\citep{yang2025qwen3}. Interaction spans multiple turns, but reward normalization and credit assignment remain trajectory-level under standard GRPO.

\section{MedDDC-Eval: Diagnosis-Decoupled Evaluation}

MedDDC-Eval combines a grounded consultation interface, a frozen shared diagnostic reader, and a fixed D/T/E evaluator. Together they turn free-form policy-elicited histories into controlled comparison objects.

\subsection{Held-Out Sources and Reference Targets}

The testbed uses two Chinese-language data sources: expert-labeled hospital records and online multi-turn consultations. The record pipeline screens 3,616 records spanning 153 source department labels into 2,904 annotated candidates across 139 normalized secondary departments. From this pool, 912 cases support RL training. Held-out evaluation is case-disjoint from training and contains a 100-case \textbf{Record split} sampled with departmental stratification and a 70-case \textbf{Dialogue split}. Dialogue cases are retained only when the primary intent is diagnostic and the source supports both diagnosis labeling and reference-trajectory construction. The Dialogue split tests whether findings transfer from record-derived cases to cases derived from online consultations. Both splits are converted into simulated consultations; the evaluated doctor never receives the retained source dialogue.

Each case is $c=(s,x,F,G,Q,b)$: data source $s$, initial context $x$, de-identified simulator facts $F$, reference diagnoses $G$, trajectory targets $Q$, and turn budget $b$. A target pairs a reference question with the information it should elicit. Annotators rate candidate targets on a three-level importance scale; only highest-importance targets enter the frozen $Q$ and contribute equally to TP/FP/FN. The target set represents normalized information needs rather than canonical wording or question order, and compatible subquestions may remain bundled as one inquiry unit. Volunteered target facts are excluded from that dialogue's trajectory denominator.

\subsection{Construction, Governance, and Simulation}

Fifteen medical annotators construct $G$ and $Q$ across four batches. The workflow proceeds from case-grounded drafting to medical normalization and importance rating, followed by review for clinical relevance, redundancy, and semantic consistency; disagreements are adjudicated before freezing. Case-level identifiers are deduplicated before split assignment, and training and evaluation are identifier-disjoint. The supplementary material reports the complete construction funnel, frozen inventory, release tiers, and validation evidence.

The grounded patient simulator answers from $F$: record cases begin from a structured complaint, whereas dialogue-derived cases use an extracted queue of retained facts. The doctor asks questions or emits a structured diagnosis action, and the rollout stops at that action or the turn budget; the main training setup permits at most seven consultation turns. This interface controls information access while allowing policies to choose different questions, orders, and stopping points.

\subsection{Shared Diagnostic Reader}

Each completed history $h_\pi$ is passed to the same frozen prompt-based DS-V3 diagnostic reader~\citep{deepseekai2025deepseekv3} $r_{\mathrm{shared}}$ with a common prompt and decoding interface. The resulting score measures diagnostic support in $h_\pi$ under a standardized history-to-diagnosis mapping. The fixed-history intervention replaces each policy's own terminal diagnosis generator with this reader. Policies are compared through elicited histories while retaining different questions and stopping points; coupled evaluation remains complementary when terminal generation is part of the target capability.

\subsection{D/T/E Measurement and One-to-One Assignment}

\begin{figure}[!tb]
\centering
\includegraphics[width=\columnwidth]{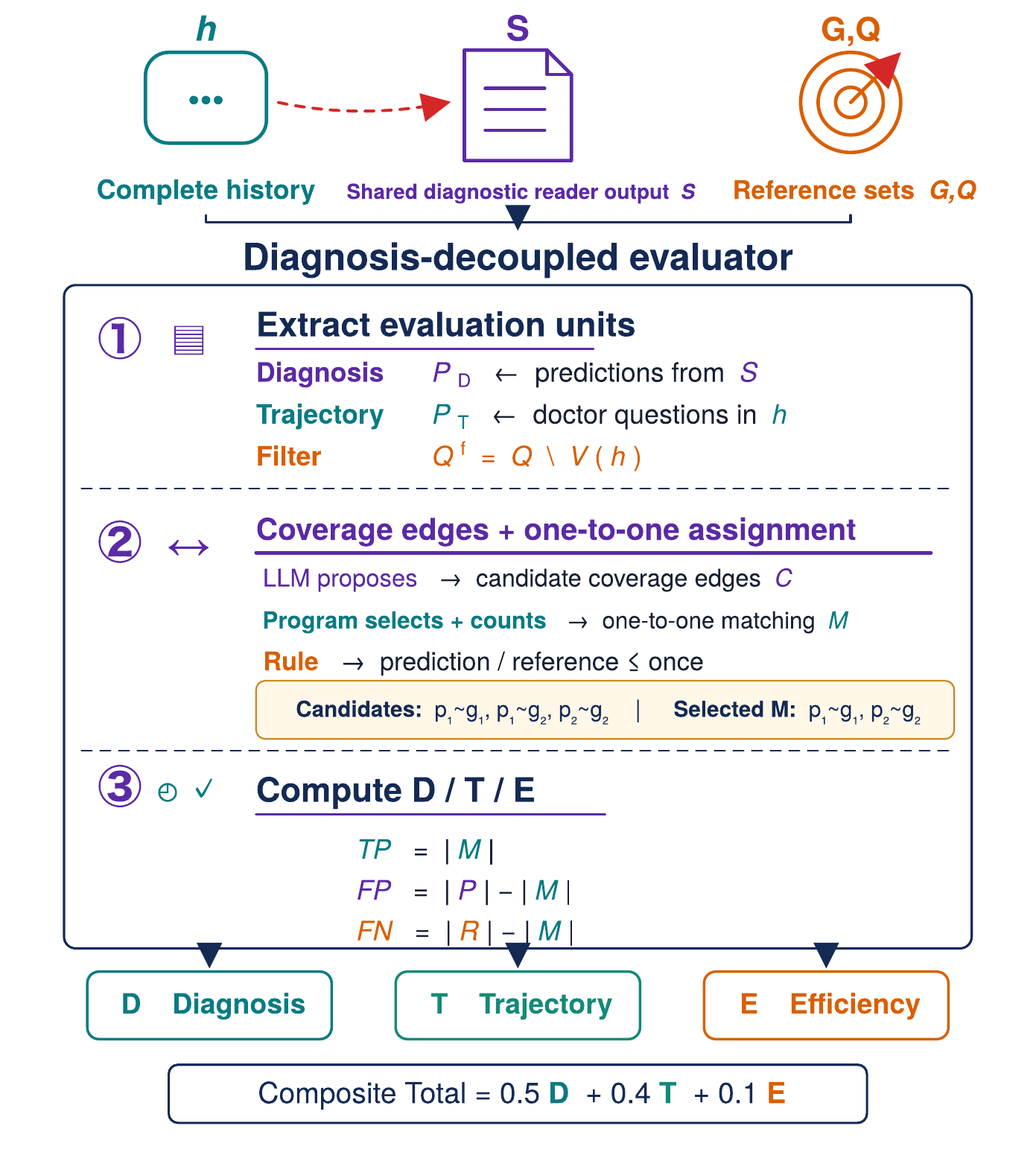}
\caption{MedDDC-Eval maps each history $h$ to diagnosis output $S$ with a frozen shared diagnostic reader and extracts diagnosis predictions and pre-summary questions. LLM judges propose semantic matches to diagnosis references $G$ and trajectory targets $Q$; deterministic one-to-one matching yields TP/FP/FN for D and T, while E captures when matched targets are covered. D/T/E are reported separately; Total gives only aggregate ordering.}
\label{fig:evaluation-protocol}
\end{figure}

The evaluator receives the complete dialogue history, the shared diagnostic reader's output, and references $(G,Q)$. It extracts diagnostic predictions and pre-summary doctor-question units, then filters trajectory targets already volunteered by the patient. Figure~\ref{fig:evaluation-protocol} summarizes the pipeline: LLM-assisted judges propose candidate semantic matches under the frozen directional-coverage rules, while a deterministic program selects one-to-one matches and computes the reported scores.

For predictions $P=\{p_i\}_{i=1}^{m}$ and references $R=\{r_j\}_{j=1}^{n}$, the semantic judge proposes a protocol-admissible candidate set $C\subseteq P\times R$. An edge records directional coverage under prespecified diagnosis-granularity or inquiry-type rules. A maximum-cardinality bipartite matching $M\subseteq C$ gives
\begin{equation}
\mathrm{TP}=|M|,\quad \mathrm{FP}=m-|M|,\quad \mathrm{FN}=n-|M|,
\end{equation}
followed by standard precision, recall, and F1. Each prediction and reference receives at most one credited match, so repeated paraphrases cannot increase TP beyond the number of distinct nodes. Candidate generation handles open-ended paraphrases and protocol-specified granularity relations, while deterministic assignment makes counting reproducible. This decomposition separates two validity questions: whether credited edges satisfy the protocol and whether candidate generation covers all scoring-relevant admissible relations. The saved candidate correspondences and their selected, excluded, and unmapped outcomes support assignment-level audit.

\textbf{Diagnosis} $D$ is run-level micro F1 between predictions from the shared diagnostic reader and $G$; it measures the diagnostic support encoded in the elicited history. \textbf{Trajectory} $T$ is run-level micro F1 between pre-summary doctor-question units and the filtered targets $Q$; it measures information-acquisition coverage against the reviewed reference-question targets while allowing flexible wording and question order. Bundling preserves clinically natural inquiry units, with one selected prediction and one selected target consuming one matching capacity. \textbf{Efficiency} $E\in[0.1,1]$ summarizes how early matched trajectory targets are covered and penalizes ineffective turns. Its full timing map appears in the supplementary material. The prespecified aggregate
\begin{equation}
\mathrm{Total}=0.5D+0.4T+0.1E
\end{equation}
provides a joint system-level ordering, while D/T/E remain visible for component-level interpretation. Tables report all four quantities on a 0--100 scale. The 100 and 70 held-out cases are the statistical units, and matching-operation counts serve as reproducibility audits. All experiments use the frozen splits, simulator, prompts, reader, matching protocol, and weights of MedDDC-Eval v1.0.

\section{Evaluation-Informed Training with Standard GRPO}

We test whether MedDDC-Eval's diagnosis and trajectory dimensions can inform policy development. Qwen3-32B~\citep{yang2025qwen3} is post-trained on cases disjoint from held-out evaluation using a separate training-time reward; the evaluator is frozen beforehand, so the study measures downstream utility under a fixed contract.

\subsection{Interactive GRPO}

Standard GRPO is applied to complete doctor--patient trajectories generated with the grounded simulator. Rollouts stop at a diagnosis action or the seven-turn limit; eight are sampled per prompt and rewards are normalized within the group. With SLIME~\citep{slime_github}, the normalized sequence reward is broadcast across policy-generated tokens. Interaction is multi-turn, while feedback and credit assignment remain trajectory-level.

\subsection{Feedback and Controls}

The reward combines diagnosis and trajectory feedback with auxiliary behavioral constraints. A training-side terminal diagnosis generator scores whether the completed history supports the target, while trajectory feedback scores annotated-target coverage. Turn, format, repetition, question-count, and cross-turn prefix-similarity terms promote valid rollouts.

Training-side generation, extraction, and semantic judging use DeepSeek-v3.2; held-out evaluation uses the frozen DS-V3 shared diagnostic reader~\citep{deepseekai2025deepseekv3} and offline $0.5D+0.4T+0.1E$ evaluator. Behavioral terms absent from the held-out score preserve separation between optimization feedback and reported measurement; coefficients and prompts appear in the supplementary material.

Ablations remove either feedback signal with the simulator, auxiliary constraints, grouped-rollout procedure, and held-out evaluator fixed.

\section{Experiments}
\label{sec:experiments}

\begin{table*}[!t]
\centering
\footnotesize
\definecolor{tablegroupgray}{gray}{0.94}
\setlength{\tabcolsep}{2.2pt}
\begin{tabular}{lrrrrrrrrc}
\toprule
Agent / policy & \multicolumn{4}{c}{Record-driven (100 cases)} & \multicolumn{4}{c}{Dialogue-driven (70 cases)} & $\Delta$ Total vs. Qwen3-32B \\
\cmidrule(lr){2-5}\cmidrule(lr){6-9}\cmidrule(lr){10-10}
 & D & T & E & Total & D & T & E & Total & Record / Dialogue \\
\midrule
\rowcolor{tablegroupgray}
\multicolumn{10}{l}{\textit{Controlled same-family comparison}} \\
Qwen3-32B + GRPO & \textbf{44.6} & 57.2 & \textbf{82.9} & \textbf{53.4} & \textbf{41.8} & \textbf{52.6} & \textbf{82.4} & \textbf{50.2} & \textbf{+9.6 / +4.6} \\
Qwen3-32B & 33.1 & 48.1 & 80.2 & 43.8 & 39.8 & 44.7 & 78.0 & 45.6 & -- \\
\rowcolor{tablegroupgray}
\multicolumn{10}{l}{\textit{External reference agents}} \\
GPT-5.4 & 36.3 & \textbf{58.8} & 79.7 & 49.7 & 35.5 & 52.4 & 79.1 & 46.6 & +5.9 / +1.0 \\
GLM-5 & 38.0 & 54.4 & 66.4 & 47.4 & 34.3 & 44.7 & 74.9 & 42.5 & +3.6 / -3.1 \\
HuatuoGPT-o1-70B & 37.7 & 46.2 & 77.9 & 45.1 & 37.2 & 40.8 & 75.9 & 42.5 & +1.3 / -3.1 \\
Llama-3-70B-UltraMedical & 31.4 & 50.5 & 81.2 & 44.0 & 39.5 & 46.9 & 79.4 & 46.4 & +0.2 / +0.8 \\
Baichuan-M2-32B & 32.3 & 46.3 & 81.8 & 42.9 & 33.5 & 44.2 & 80.8 & 42.5 & -0.9 / -3.1 \\
Doctor-R1 & 31.5 & 40.3 & 75.0 & 39.3 & 37.4 & 41.2 & 78.6 & 43.1 & -4.5 / -2.5 \\
\bottomrule
\end{tabular}
\caption{Main results under the shared diagnostic reader protocol. Values are percentage means over five fixed-checkpoint rollout repeats; within each repeat, D and T use micro-aggregated one-to-one counts and E uses the first summary round. Qwen3-32B + GRPO versus Qwen3-32B is the controlled training comparison; external agents provide descriptive context. Total is $0.5D+0.4T+0.1E$, and $\Delta$ is computed from the displayed one-decimal Total values for Record / Dialogue. Sample standard deviations appear in the supplementary material.}
\label{tab:main-results-compact}
\end{table*}

\subsection{Experimental Logic and Common Evaluation Setting}

The experiments proceed in three stages. We first isolate whether the terminal diagnosis component changes comparisons among fixed consultation histories. We then test whether the evaluation dimensions support controlled policy development and whether diagnosis and trajectory feedback make distinct contributions. Finally, we characterize case-level heterogeneity and sensitivity to rollout repeats, score weights, and alternative readers\ifdefined\ARXIVPREPRINT\else, and use physician review to examine credited-match validity and component-level alignment\fi.

The main comparison contains the trained policy, its Qwen3-32B initialization, and six external consultation agents listed in Table~\ref{tab:main-results-compact}~\citep{yang2025qwen3,openai2026gpt54,glm5team2026glm5,chen2024huatuogpto1,zhang2024ultramedical,m2team2025baichuanm2,lai2025doctorr1}. Every agent uses the same Chinese-language simulator, source-specific turn budget, frozen DS-V3 shared diagnostic reader, extraction and matching protocol, D/T/E weights, and five-run reporting procedure. We use the trained--base pair as the controlled training contrast; external systems provide reference context under the common evaluation protocol. Repeats sample new rollouts from fixed checkpoints, while confidence intervals resample held-out cases as clusters and recompute the run-level estimators.

\subsection{Reader Control Exposes Attribution Confounding}

The terminal diagnosis component materially changes between-policy comparisons even with fixed consultation histories. Figure~\ref{fig:object-vector} holds each history, diagnostic prompt, and decoding setting fixed while replacing each policy's own terminal diagnosis generator with the shared diagnostic reader. Diagnosis F1 shifts by 2.2--19.0 points, reversing 5/28 (18\%) Record and 10/28 (36\%) Dialogue pairwise orderings in the complete eight-policy audit. Questions, patient answers, trajectory, and efficiency remain fixed, so the heterogeneous shifts and reversals identify terminal diagnosis generation as the source of the comparison change and motivate use of the frozen shared diagnostic reader.

Diagnosis and trajectory are related across systems but are not interchangeable for individual consultations. Across eight system means, Pearson $r$ is 0.566 on Record and 0.303 on Dialogue; across case--model means it falls to 0.147 and 0.101. High-trajectory/low-diagnosis and low-trajectory/high-diagnosis cases both remain common. The axes can improve together across systems, but neither reliably substitutes for the other on individual consultations; D/T component profiles therefore carry information hidden by Total.

\subsection{Evaluation-Informed Training Improves Elicited Histories}

A training reward informed by the diagnosis and trajectory dimensions improves both diagnostic support and information-acquisition coverage. In the controlled trained--base comparison, the architecture family and initialization are shared, so the contrast isolates the effect of post-training under the reported setup. The six external consultation agents in Table~\ref{tab:main-results-compact} provide descriptive context under the common evaluation protocol.

Relative to its initialization, the trained policy improves Record D/T/E/Total by 11.5/9.1/2.7/9.6 points and Dialogue by 2.0/7.9/4.4/4.6 points. The trained policy has the highest Total in the eight-system comparison, while GPT-5.4 retains the strongest Record trajectory score by 1.6 points. This contrast demonstrates why component profiles complement the aggregate ranking.

These profiles change how benchmark results should be used. Diagnosis F1 is the direct comparison when downstream diagnostic support is the priority, whereas trajectory F1 indicates whether the policy elicited the reviewed information needs. Efficiency is most informative after D and T reach an acceptable level: a fast but incomplete consultation can otherwise appear favorable. Models with similar Total can therefore suit different uses, and reporting only the aggregate would hide trade-offs that matter for model selection.

A case-cluster bootstrap over the cases available for each comparison supports the headline mean gains. Each resample re-aggregates TP/pred/gold within every run, recomputes micro D/T, case-mean E, and Total, and then averages the five repeats. The resulting trained--base Total differences are 9.6 points on Record (95\% CI [7.6, 11.8]) and 4.6 on Dialogue ([2.4, 6.8]); differences from GPT-5.4 are 3.8 ([1.4, 6.3]) and 3.6 ([1.0, 6.2]). These intervals quantify held-out case-sampling uncertainty for the evaluated checkpoints; training-seed variation lies outside this fixed-checkpoint analysis.

\subsection{Case-Level Gains Are Heterogeneous}

\begin{figure}[!t]
\centering
\includegraphics[width=\columnwidth]{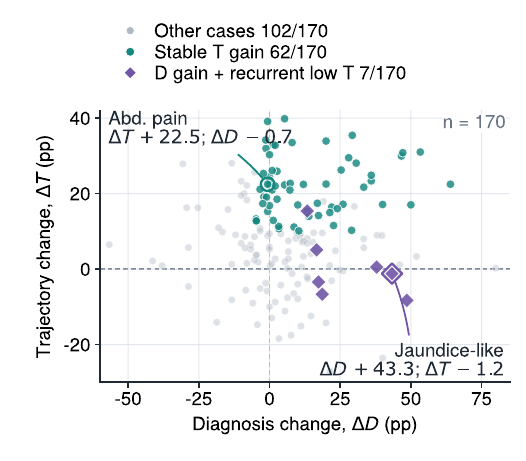}
\caption{Case-level trained--base changes under the shared diagnostic reader ($n=170$; five-run means). $S$ is the fraction of the 25 trained--base cross-run pairs in which the trained policy scores higher. S2 (teal; 62/170) requires $\Delta T\geq10$, $S_T\geq0.70$, and $\Delta D\geq-5$. S3 (violet; 7/170) requires $\Delta D\geq10$, $S_D\geq0.70$, and split-specific bottom-$T$/top-$D$ membership in at least 3/5 trained-policy runs. For S3, low $T$ denotes an absolute level, not $\Delta T<0$. Patterns are predefined, non-exhaustive, and may overlap.}
\label{fig:case-delta-decomposition}
\end{figure}

Aggregate gains conceal heterogeneous case-level improvement modes. Figure~\ref{fig:case-delta-decomposition} decomposes the trained--base effect over 170 complete case pairs, using differences between five-run case means and the cross-run stability criteria in the caption. Sixty-two cases (36.5\%) meet the S2 stable trajectory-gain criterion; the annotated abdominal-pain case lies near this bucket's center with trajectory $+22.5$ points and diagnosis $-0.7$. Seven cases (4.1\%) meet the S3 diagnosis-gain criterion with recurrently sparse absolute trajectory coverage; the annotated jaundice-like case gains $43.3$ diagnosis points while trajectory changes by $-1.2$.

The two modes also support targeted case analysis. High-trajectory/low-diagnosis cases identify histories that cover reviewed questions yet provide limited diagnostic support under the shared diagnostic reader. Low-trajectory/high-diagnosis cases show that salient clues can support a diagnosis despite limited information-acquisition coverage. These buckets can prioritize case review and indicate whether future reward revisions should emphasize target coverage or diagnostic support; their role is to localize observed behavior for follow-up analysis.

\subsection{Both Feedback Signals Contribute}

\begin{table}[!t]
\centering
\footnotesize
\setlength{\tabcolsep}{2.5pt}
\begin{tabular}{lrrrr}
\toprule
Setting & D & T & E & Total \\
\midrule
\multicolumn{5}{l}{\textit{Record-driven}} \\
Full reward & \textbf{44.6} & \textbf{57.2} & \textbf{82.9} & \textbf{53.4} \\
w/o diagnosis feedback & 36.4 & 53.2 & 90.8 & 48.5 \\
w/o trajectory feedback & 38.8 & 55.4 & 78.1 & 49.4 \\
\addlinespace[2pt]
\multicolumn{5}{l}{\textit{Dialogue-driven}} \\
Full reward & \textbf{41.8} & \textbf{52.6} & \textbf{82.4} & \textbf{50.2} \\
w/o diagnosis feedback & 35.3 & 50.1 & 89.0 & 46.6 \\
w/o trajectory feedback & 36.3 & 50.3 & 77.0 & 46.0 \\
\bottomrule
\end{tabular}
\caption{Reward-component ablations for Qwen3-32B policies trained with GRPO. Values are percentage means over five fixed-checkpoint rollout repeats; the full-reward rows match the trained policy in Table~\ref{tab:main-results-compact}. Ablations change only the training reward, while the held-out evaluator and $0.5D+0.4T+0.1E$ score remain fixed.}
\label{tab:reward-ablations}
\end{table}

Both feedback signals provide non-redundant gains in this controlled Qwen3-32B setting. Removing diagnosis feedback lowers Total by 4.9/3.6 points on Record/Dialogue, while removing trajectory feedback lowers it by 4.1/4.2 points. Case-cluster bootstrap confidence intervals for all four differences remain above zero. With the simulator, auxiliary constraints, grouped-rollout procedure, and held-out evaluator fixed, the removals are directly comparable. The no-diagnosis variant nevertheless increases E while reducing D and T, showing why efficiency must be interpreted jointly.

\ifdefined\ARXIVPREPRINT
\subsection{Evaluator Stability Under the Fixed Protocol}
\else
\subsection{Meta-Evaluation Separates Stability and Alignment}
\fi

The main ordering is stable across the tested evaluator perturbations, although separation narrows in the most sensitive setting. Across the 16 model--split summaries, the median Total SD over five rollout repeats is 1.41 points, and the top mean ordering is preserved across all seven tested weight settings and both alternative readers. Under the GPT-5.4 Dialogue reader, however, the leading margin narrows to 0.44 points; this result supports ordering stability in that setting, but not decisive pairwise separation.

\ifdefined\ARXIVPREPRINT
These checks characterize sensitivity to rollout randomness, aggregate weights, and the history-to-diagnosis reader. They do not establish prospective clinical validity or exhaustive coverage of all admissible semantic matches; those questions remain outside this offline benchmark.
\else
Physician review examines a different property: the validity and relative alignment of the protocol-defined comparisons. Two physicians independently reviewed 180 selected semantic matches, reaching 93.3\% raw agreement; after adjudication, 174/180 matches awarded credit were protocol-valid (96.7\%, Wilson 95\% CI [92.9\%, 98.5\%]), including 78/80 diagnosis and 96/100 trajectory matches. In a separate audit, six medically qualified reviewers independently compared four anonymized systems on each of ten cases along diagnostic support, information-acquisition coverage, and efficiency. Majority judgments formed the physician consensus, with ties or substantive disagreements resolved by a senior physician. Automatic D/T/E comparisons agreed with physician-consensus pairwise preferences in 166/180 comparisons (92.2\%, Wilson 95\% CI [87.4\%, 95.3\%]). These audits support the validity of sampled credited matches and the physician alignment of protocol-defined relative comparisons; full protocols and scope boundaries appear in the supplementary material.
\fi

\FloatBarrier

\section{Limitations}

MedDDC-Eval evaluates simulated Chinese-language consultations rather than clinical deployment. Grounded simulators provide controlled, case-consistent interactions but do not capture the full diversity of patient behavior, and valid consultations may differ from the reviewed reference trajectories in ordering and phrasing.

Results are conditional on the frozen shared diagnostic reader and candidate-generation protocol. \ifdefined\ARXIVPREPRINT Broader clinical and cross-setting validation, as well as independent training-seed studies, remain future work.\else Two physician audits support the validity of sampled matches awarded credit and the relative D/T/E comparisons in the audited cases, while broader clinical and cross-setting validation remains future work.\fi\ The reward-ablation study evaluates one Qwen3-32B training configuration with repeated rollouts from fixed checkpoints; its conclusions therefore apply to the reported controlled setting.

\section{Ethical Statement}

This work uses de-identified source-derived cases to drive simulated consultations for offline evaluation. Experimental inputs contain no direct personal identifiers; raw records, original dialogues, linkable identifiers, and case-level traces are not released, as detailed in the supplementary data-construction and validation appendix. No evaluated model is intended for patient-facing or autonomous clinical use, and automatic scores are not safety certification.

Planned public artifacts are limited to reviewed prompts, schema, scoring code, aggregate results, and synthetic examples. Source-derived subsets remain conditional on authorization, licensing, institutional review, and de-identification; these constraints take precedence over unrestricted release.

\section{Conclusion}

MedDDC-Eval reframes the evaluation of multi-turn medical consultation agents around the histories that policies elicit. A frozen shared diagnostic reader holds the history-to-diagnosis mapping constant across policies, while one-to-one D/T/E measurement characterizes diagnostic support, information-acquisition coverage, and efficiency. Fixed-history interventions reveal score shifts and ranking reversals when policy-specific terminal diagnosis generators are replaced, and the controlled GRPO study shows gains in held-out elicited-history measurements using a separate reward organized around diagnosis and trajectory feedback. Together, these results position diagnosis-decoupled evaluation as a complement to end-to-end evaluation: the former compares elicited histories under a common reader, whereas the latter assesses the complete system, including terminal diagnosis generation. Future work should examine broader reader ensembles, adaptive diagnosis--trajectory feedback, and additional language and evaluation settings.

\bibliographystyle{abbrvnat}
\bibliography{custom}

\clearpage
\onecolumn
\raggedbottom
\makeatletter
\setlength{\@fptop}{0pt}
\setlength{\@fpsep}{12pt}
\setlength{\@fpbot}{0pt plus 1fil}
\setlength{\@dblfptop}{0pt}
\setlength{\@dblfpsep}{12pt}
\setlength{\@dblfpbot}{0pt plus 1fil}
\makeatother
\renewcommand{\topfraction}{0.95}
\renewcommand{\bottomfraction}{0.90}
\renewcommand{\textfraction}{0.06}
\renewcommand{\floatpagefraction}{0.78}
\setlength{\textfloatsep}{14pt plus 2pt minus 2pt}
\setlength{\floatsep}{12pt plus 2pt minus 2pt}
\setlength{\dbltextfloatsep}{14pt plus 2pt minus 2pt}
\setlength{\dblfloatsep}{12pt plus 2pt minus 2pt}
\renewcommand{\arraystretch}{1.04}
\makeatletter
\renewenvironment{table*}[1][]{\@float{table}[!htbp]}{\end@float}
\makeatother
\newcommand{\supplementfigurewidth}{0.58\textwidth}
\section{Supplementary Material}
\appendix
\setcounter{secnumdepth}{3}
\renewcommand{\thesection}{\Alph{section}}
\renewcommand{\thesubsection}{\thesection.\arabic{subsection}}
\renewcommand{\thesubparagraph}{\thesubsection.\arabic{subparagraph}}
\makeatletter
\renewcommand{\section}[1]{%
  \FloatBarrier
  \par\vspace{2.0ex plus 0.5ex minus 0.2ex}%
  \refstepcounter{section}%
  \noindent\parbox{\columnwidth}{\centering\Large\bfseries Appendix~\thesection\\#1}\par%
  \vspace{3pt plus 2pt minus 1pt}%
  \nopagebreak[4]%
}
\makeatother
\renewcommand{\thetable}{S\arabic{table}}
\renewcommand{\thefigure}{S\arabic{figure}}
\setcounter{table}{0}
\setcounter{figure}{0}

\section{Dataset Construction and Validation}
\label{sec:dataset-construction-validation}

\subsection{Sources, Funnel, and Experimental Language}

All source records, online consultation dialogues, simulator interactions, doctor outputs, shared diagnostic reader inputs, and evaluator-facing outputs are Chinese. No translation is used in the experimental pipeline. The authoritative runtime prompts are Chinese; English translations are explanatory, non-executable reading aids.

Table~\ref{tab:dataset-stats} traces the record-source construction funnel and the two MedDDC-Eval held-out splits. The 912 training cases support the evaluation-informed training study and are instantiated once in each of the 3--4 and 5--7 turn buckets. The 100-case Record split is drawn from the same annotated candidate pool with departmental stratification, and the 70-case Dialogue split is constructed from online-consultation sources with frozen diagnosis-label and reference-trajectory inventories. Training and held-out evaluation are separated by case identifier. Semantic near-duplicate screening across the heterogeneous source text remains a follow-up to the completed identifier-level deduplication.

\subsection{Case Schema and Reference Construction}

The common case object contains a data-source label, initial patient-facing context, a de-identified simulator fact store, reference diagnoses, trajectory targets, and a turn budget. Table~\ref{tab:annotation-schema} specifies how these fields enter simulation and evaluation. A trajectory target pairs a reference question with its associated clinical condition or fact; it denotes a normalized clinical information need rather than a literal gold wording or a unique path. Targets may retain related subquestions as clinically natural bundles, and semantic matching under directional coverage rules separates information-acquisition coverage from wording imitation. Volunteered-target status is computed from each realized dialogue rather than treated as a static case label.

Fifteen medical annotators worked across four batches. They annotated targets judged to bear on the diagnosis or treatment plan. Construction followed three stages: case-grounded drafting; medical normalization with a three-level importance rating, of which only the highest-rated targets are retained; and review of clinical relevance, redundancy, and semantic consistency with adjudication before freezing. This workflow produces a reviewed consensus reference rather than a matrix of independent parallel labels; construction validity is therefore assessed through staged medical review.

The staged review above provides construction validity by defining trajectory targets as reviewed normative information needs while preserving flexibility in consultation sequence. Scoring integrity is addressed through the evaluator contract and sensitivity analyses below.

\begin{table*}[t]
\centering
\footnotesize
\textbf{(a) Record-source construction funnel and dataset roles}\\[3pt]
\setlength{\tabcolsep}{3.5pt}
\resizebox{0.98\textwidth}{!}{%
\begin{tabular}{lrrll}
\toprule
Construction output & Cases & Prompt rows & Scope & Role \\
\midrule
Initial record pool & 3,616 & -- & 153 source department labels & Construction input \\
Eligible annotated candidates & 2,904 & -- & 139 normalized secondary departments & Candidate pool \\
Curated RL cases & 912 & 1,824 & 3--4 and 5--7 turn buckets & Main RL training \\
Scaled RL candidates & 2,904 & 5,808 & Same two turn buckets & Exploratory scaling only \\
Record-driven evaluation & 100 & 100 & Held-out record-derived cases & Main evaluation \\
Dialogue-driven evaluation & 70 & 70 & Held-out dialogue-derived cases & Source-shift evaluation \\
\bottomrule
\end{tabular}
}

\vspace{6pt}
\textbf{(b) Frozen held-out reference inventory}\\[3pt]
\begin{tabular}{lrrrr}
\toprule
Stratum & Cases & Diagnosis labels & Trajectory targets & Targets/case, median (range) \\
\midrule
Record-driven & 100 & 190 & 863 & 8 (5--18) \\
Dialogue-driven & 70 & 97 & 588 & 8 (4--14) \\
\midrule
Total & 170 & 287 & 1,451 & -- \\
\bottomrule
\end{tabular}
\caption{Non-identifying dataset construction and held-out reference statistics. A training case yields two prompt rows because it is instantiated under two turn-budget buckets. Held-out labels and targets are counted once per unique case before consultation-specific volunteered-fact filtering. For the Dialogue split, online-consultation sources are converted into the same two frozen reference inventories: diagnosis labels and trajectory targets. Counts describe construction outputs rather than annotator-hours, population prevalence, or sampling rates. Raw records and original dialogues are not released.}
\label{tab:dataset-stats}
\end{table*}

\begin{table*}[t]
\centering
\footnotesize
\setlength{\tabcolsep}{3.5pt}
\begin{tabular}{p{2.6cm}p{4.8cm}p{6.7cm}}
\toprule
Case field & Construction meaning & Simulation, scoring, and release role \\
\midrule
Data source $s$ & Record-derived or dialogue-derived provenance. & Defines the evaluation split; only aggregate source labels are public. \\
Initial context $x$ & Patient-facing complaint or retained dialogue context used to start the consultation. & Text is released only in synthetic or approved fully anonymized examples. \\
Grounded fact store $F$ & De-identified case facts available to the patient simulator. & Constrains patient replies; patient-derived source text and linkages are not released. \\
Diagnosis set $G$ & Reference diagnostic labels supported by the case. & Used by one-to-one diagnosis scoring; labels are public only in synthetic or approved controlled instances. \\
Trajectory target $q$ & Normalized clinical information need the doctor should elicit, not a verbatim source span or unique gold wording. & Scored through semantic matching under directional coverage and inquiry-type constraints. \\
Reference question & Case-grounded natural-language question associated with the trajectory target. & Provides the reference node for coverage without requiring the model to reproduce its wording. \\
Clinical condition/fact & Case fact or condition that the reference question is intended to elicit. & Grounds interpretation of the target while protected source text and linkages remain unreleased. \\
Importance rating & Three-level construction-time rating of how strongly a drafted target bears on the diagnosis or treatment plan. & Only highest-rated targets enter the frozen reference set; each frozen target counts once in one-to-one TP/FP/FN. \\
Volunteered status & Dialogue-specific indicator that the patient supplied the target before it was asked. & Derived after rollout and filtered from that dialogue's denominator; not a static gold label. \\
Turn budget $b$ & Maximum consultation horizon for the instantiated prompt. & Controls rollout length; training cases use the 3--4 and 5--7 buckets. \\
\bottomrule
\end{tabular}
\caption{Common case and annotation schema. Reference trajectories are sets of reviewed trajectory targets, not a single literal sequence. Related subquestions may remain bundled as a natural inquiry unit; one-to-one assignment limits each predicted and reference unit to one credited edge.}
\label{tab:annotation-schema}
\end{table*}

\section{Evaluator Contract and One-to-One Assignment}

\subsection{Matching and Aggregation}

The evaluator extracts diagnostic items and clinically meaningful doctor-question units, obtains candidate semantic matches from the LLM-assisted judge under the protocol's directional coverage rules, and applies deterministic maximum-cardinality bipartite matching. If $m$ predictions and $n$ references yield a matching of size $k$, then $\mathrm{TP}=k$, $\mathrm{FP}=m-k$, and $\mathrm{FN}=n-k$. Candidate edges after the first summary action are excluded from trajectory scoring, and reference items already volunteered by the patient are filtered before matching.

The saved candidate correspondences record the candidate, selected, excluded, and unmapped outcomes used for counting. The proposal set follows the frozen protocol, and MedDDC-Eval v1.0 micro-aggregates the resulting counts within each run before reporting the mean and sample SD across five fixed-checkpoint rollout repeats. The held-out shared diagnostic reader, extractor, and candidate-match judge use DeepSeek-v3 through frozen role-specific gateways; the training-side terminal diagnosis generator, extractor, and judge use DeepSeek-v3.2. The original Chinese prompts are authoritative runtime artifacts.

\begin{table*}[t]
\centering
\small
\setlength{\tabcolsep}{3pt}
\begin{tabular}{p{2.8cm}p{3.2cm}p{3.2cm}p{5.2cm}}
\toprule
Evaluator step & LLM / parameters & Expected output & Fallback or validation \\
\midrule
Medical item extraction & DeepSeek-v3; temperature 0.2, top-p 0.9, JSON mode when used. & Dimension-specific top-k lists for diagnosis, tests, treatment, medication, lifestyle advice, and medical guidance. & Structured tool-call content is used first when available; otherwise the evaluator extracts from the full dialogue. Invalid or missing fields are normalized to empty lists. \\
Dimension coverage judgment & Configured dimension-judgment prompt with the same default evaluator model. & Directional, protocol-admissible coverage edges between extracted predictions and references. & Parsed JSON is structure-validated; malformed items produce no candidate edge. A deterministic maximum-cardinality matcher selects the final one-to-one edges. \\
Question extraction & DeepSeek-v3 with the configured trajectory-question-extraction prompt; JSON object response. & List of doctor questions extracted from the dialogue. & If LLM extraction fails, the extracted list is empty and downstream trajectory matching receives no matched questions. \\
Question-to-turn alignment & String containment, then LLM semantic match, then keyword containment. & Round index for each extracted doctor question. & Semantic match uses the trajectory-matching prompt; failure falls back to keyword containment only when a sufficiently long question string is available. \\
Trajectory coverage matching & Configured trajectory-matching prompt; JSON object response. & Candidate question--target coverage edges with aligned rounds. & Bundled targets permit valid component coverage under type constraints; post-summary and invalid edges are discarded; deterministic one-to-one matching produces the TP/FP/FN counts. \\
Patient-volunteered filtering & DeepSeek-v3 with patient-coverage prompt; temperature 0.0, top-p 0.9, JSON mode. & Whether an unmatched gold target was already volunteered by the patient. & If model judgment is unavailable, string fallback checks whether the question or condition text appears in patient turns. \\
Efficiency computation & Deterministic after matched rounds are available. & Timing score from matched trajectory rounds and ineffective-question penalty. & No additional LLM call is used for the timing formula. \\
\bottomrule
\end{tabular}
\caption{LLM-assisted candidate generation followed by deterministic one-to-one assignment. The LLM proposes semantic edges; programmatic matching determines the final confusion counts.}
\label{tab:llm-judge-protocol}
\end{table*}

\subsection{Efficiency}

Let $\mathcal{T}=\{t_1,\ldots,t_m\}$ contain the first matched rounds before the first summary action. If no item is matched, $E=0.1$. Otherwise, with $\lambda=0.15$,
\begin{equation}
w_i=\exp[-\lambda(t_i-1)],\qquad \bar{t}=\frac{\sum_i t_iw_i}{\sum_i w_i}.
\end{equation}
The timing map is
\begin{equation}
\mathrm{timing}(\bar{t})=
\begin{cases}
1.0,&\bar{t}\leq2,\\
0.95-0.05(\bar{t}-2),&2<\bar{t}\leq3,\\
0.90-0.15(\bar{t}-3),&3<\bar{t}\leq5,\\
0.60-0.10(\bar{t}-5),&5<\bar{t}\leq8,\\
\max(0.1,0.30-0.02(\bar{t}-8)),&\bar{t}>8.
\end{cases}
\end{equation}
With $T$ the first summary round, or exported total rounds when no summary is found,
\begin{align}
r_{\mathrm{ineff}}&=\max\left(0,\frac{T-m}{T}\right),\\
E&=\min\left(1,\max\left(0.1,\mathrm{timing}(\bar{t})\,
\bigl(1-0.3r_{\mathrm{ineff}}\bigr)\right)\right).
\end{align}

\ifdefined\ARXIVPREPRINT
\else
\section{Evaluator Meta-Evaluation and Measurement Boundaries}

The meta-evaluation combines fixed-checkpoint repeats, local aggregate-weight perturbations, alternative shared diagnostic readers, direct physician review of semantic matches awarded credit, and a blinded D/T/E physician-consensus audit. For match validity, two physicians independently reviewed 180 selected matches: 80 diagnosis and 100 trajectory assignments. Raw agreement was 93.75\% for diagnosis, 93.00\% for trajectory, and 93.33\% overall. After adjudication, 78/80 diagnosis matches and 96/100 trajectory matches were protocol-valid, yielding 174/180 overall (96.67\%, Wilson 95\% CI [92.92\%, 98.46\%]). This audit evaluates sampled matches awarded credit; exhaustive candidate-generation coverage remains a separate measurement question. In the component-level audit, six medically qualified reviewers assessed four anonymized system outputs for each of ten cases while blinded to system identity, training configuration, and automatic scores. They independently compared diagnostic support, information-acquisition coverage, and efficiency; majority judgments formed the physician consensus, with ties or substantive disagreements resolved through senior-physician adjudication. Automatic D/T/E comparisons agreed with physician-consensus pairwise preferences in 166/180 comparisons (92.2\%, Wilson 95\% CI [87.4\%, 95.3\%]). This audit evaluates physician alignment of protocol-defined relative comparisons; continuous-score calibration and prospective clinical validity require separate study designs.

\input{tables/evaluator_meta_evaluation}
\input{tables/physician_selected_edge_audit}
\input{tables/physician_eval_protocol}
\fi

\section{Fixed-History Diagnostic-Reader Intervention}

Supplementary Table~\ref{tab:supp-reader-intervention} provides the complete eight-policy fixed-history audit underlying the main-paper intervention figure. Each rollout history and the diagnostic prompt are held fixed; only the terminal diagnosis component changes from each policy's own terminal diagnosis generator to the frozen DS-V3 shared diagnostic reader. The table retains all exact diagnosis-F1 endpoints and rank changes, including the trained policy excluded from the figure's seven-policy aggregate braid.

\begin{table*}[t]
\centering
\scriptsize
\setlength{\tabcolsep}{2pt}
\begin{tabular}{@{}lcccc@{}}
\toprule
Policy rollout & Record diagnosis F1 & Record D rank & Dialogue diagnosis F1 & Dialogue D rank \\
 & \shortstack{Own generator $\rightarrow$\\shared diagnostic reader} & \shortstack{Own generator $\rightarrow$\\shared diagnostic reader} & \shortstack{Own generator $\rightarrow$\\shared diagnostic reader} & \shortstack{Own generator $\rightarrow$\\shared diagnostic reader} \\
\midrule
Qwen3-32B + GRPO & 40.4 $\rightarrow$ 44.6 {\scriptsize(+4.1)} & 1 $\rightarrow$ 1 & 36.5 $\rightarrow$ 41.8 {\scriptsize(+5.3)} & 1 $\rightarrow$ 1 \\
Qwen3-32B & 26.5 $\rightarrow$ 33.1 {\scriptsize(+6.6)} & 6 $\rightarrow$ 5 & 29.6 $\rightarrow$ 39.8 {\scriptsize(+10.2)} & 5 $\rightarrow$ 2 \\
GPT-5.4 & 33.2 $\rightarrow$ 36.3 {\scriptsize(+3.1)} & 3 $\rightarrow$ 4 & 30.9 $\rightarrow$ 35.5 {\scriptsize(+4.6)} & 3 $\rightarrow$ 6 \\
GLM-5 & 33.9 $\rightarrow$ 38.0 {\scriptsize(+4.1)} & 2 $\rightarrow$ 2 & 30.5 $\rightarrow$ 34.3 {\scriptsize(+3.8)} & 4 $\rightarrow$ 7 \\
HuatuoGPT-o1-70B & 33.0 $\rightarrow$ 37.7 {\scriptsize(+4.7)} & 4 $\rightarrow$ 3 & 32.4 $\rightarrow$ 37.2 {\scriptsize(+4.7)} & 2 $\rightarrow$ 5 \\
Llama-3-70B-UltraMedical & 29.2 $\rightarrow$ 31.4 {\scriptsize(+2.2)} & 5 $\rightarrow$ 8 & 20.8 $\rightarrow$ 39.5 {\scriptsize(+18.6)} & 7 $\rightarrow$ 3 \\
Baichuan-M2-32B & 16.6 $\rightarrow$ 32.3 {\scriptsize(+15.8)} & 8 $\rightarrow$ 6 & 14.5 $\rightarrow$ 33.5 {\scriptsize(+19.0)} & 8 $\rightarrow$ 8 \\
Doctor-R1 & 24.9 $\rightarrow$ 31.5 {\scriptsize(+6.6)} & 7 $\rightarrow$ 7 & 27.8 $\rightarrow$ 37.4 {\scriptsize(+9.6)} & 6 $\rightarrow$ 4 \\
\bottomrule
\end{tabular}
\caption{Fixed-history terminal-diagnosis intervention. The own-generator condition applies each policy's own terminal diagnosis generator under a common diagnostic prompt; the shared-reader condition applies the frozen DS-V3 diagnostic reader to the identical history. Questions, patient answers, trajectory score, and efficiency are held fixed. The heterogeneous diagnosis-F1 shifts and rank changes show that the terminal diagnosis component materially affects diagnosis-based comparisons of elicited histories.}
\label{tab:supp-reader-intervention}
\end{table*}

\section{Compact Measurement Summaries}

Tables~\ref{tab:medddc-component-map}--\ref{tab:metric-contract} summarize the compact inventory and measurement-contract details that support the main-paper protocol description.

\begin{table*}[t]
\centering
\footnotesize
\setlength{\tabcolsep}{4pt}
\begin{tabular}{p{2.9cm}p{4.0cm}p{4.2cm}p{4.0cm}}
\toprule
Component & Role & Inventory / scope & Boundary \\
\midrule
Grounded consultation interface & Produces bounded policy-elicited histories from two data sources. & 170 unique held-out cases: 100 Record and 70 Dialogue; grounded simulator; first-summary boundary. & Offline simulated-consultation setting with 170 held-out clinical cases. \\
Shared diagnostic reader & Holds terminal diagnosis generation constant across policies. & One frozen DS-V3 prompt/decoding interface applied to every completed history. & Between-agent attribution control; scores remain conditional on the reader. \\
Auditable D/T/E harness & Measures diagnostic support, information-acquisition coverage, and efficiency. & 287 diagnosis labels; 1,451 trajectory targets; semantic candidate edges plus deterministic one-to-one assignment. & D/T/E are primary axes; Total is an aggregate summary. \\
Repeated evaluation workload & Characterizes rollout variability and supports audit. & Main-8 $\times$ 5 runs; 6,800 prespecified case--system--run evaluation slots. & The clinical sample remains the 170 held-out cases. \\
\bottomrule
\end{tabular}
\caption{MedDDC-Eval component and inventory map. The held-out clinical sample, frozen reference inventory, and repeated evaluation workload are different units and are reported separately.}
\label{tab:medddc-component-map}
\end{table*}

\begin{table*}[t]
\centering
\footnotesize
\setlength{\tabcolsep}{3.5pt}
\begin{tabular}{p{1.0cm}p{4.5cm}p{2.8cm}p{3.2cm}p{3.6cm}}
\toprule
$M$ & Question answered & Evaluation unit & Aggregation & Main boundary \\
\midrule
$D$ & Does the elicited history support reference diagnoses under the shared diagnostic reader? & Diagnosis item & Run-level micro F1 & Conditional on the frozen reader and proposed semantic edges. \\
$T$ & Does the consultation cover the reviewed reference-question targets? & Doctor question--target unit & Run-level micro F1 before first summary & Bundled-target resolution; volunteered facts are filtered. \\
$E$ & How early are matched trajectory targets covered? & Matched question turn & Deterministic timing score & Coupled to targets matched for $T$; auxiliary axis. \\
Total & What is the prespecified joint ordering? & System--run & $0.5D+0.4T+0.1E$ & Prespecified aggregate summary for system comparison. \\
\bottomrule
\end{tabular}
\caption{Self-contained D/T/E measurement contract. Semantic models propose protocol-admissible candidate edges; deterministic one-to-one assignment selects and counts them.}
\label{tab:metric-contract}
\end{table*}

\section{Evaluation-Informed Training Details}

Tables~\ref{tab:training-config}--\ref{tab:multi-turn-grpo-contract} and Figure~\ref{fig:process-aware-reward} specify the rollout, reward, and update controls used for the evaluation-informed training study.

\begin{table*}[t]
\centering
\small
\begin{tabular}{ll}
\toprule
Configuration & Group Relative Policy Optimization (GRPO) setting \\
\midrule
Base family & Qwen3-32B \\
Initialization & Qwen3-32B checkpoint \\
Rollout function & Custom multi-turn rollout \\
Reward function & Process-aware reward including diagnosis feedback and trajectory feedback \\
Advantage estimator & GRPO \\
Maximum turns & 7 \\
Samples per prompt & 8 \\
Rollout batch size & 32 \\
Rollouts & 300 \\
Maximum response length & 2,048 tokens \\
Sampling temperature & 1.0 \\
Global batch size & 64 \\
KL loss coefficient & 0.005 \\
Clipping & $\epsilon=0.2$, high clip $0.28$ \\
Optimizer & Adam, learning rate $1{\times}10^{-6}$, weight decay 0.01 \\
Parallelism & 1 node, 8 actor GPUs, tensor parallelism 4 \\
\bottomrule
\end{tabular}
\caption{Process-aware GRPO training configuration.}
\label{tab:training-config}
\end{table*}

\subsection{Rollout, Reward, and Ablation Controls}

Table~\ref{tab:multi-turn-grpo-contract} summarizes the interaction unit, termination rule, grouped sampling, reward timing, credit scope, optimizer, and train--evaluation separation used in the evaluation-informed training study.

\begin{table*}[t]
\centering
\footnotesize
\setlength{\tabcolsep}{4pt}
\begin{tabular}{p{3.1cm}p{7.0cm}p{4.7cm}}
\toprule
Field & Reported setting & Interpretive boundary \\
\midrule
Training unit / interaction & Complete doctor--patient trajectory; policy alternates with the grounded simulator. & Multi-turn structure is introduced through environment interaction. \\
Termination & Structured diagnosis action or seven-turn limit. & Defines the trajectory receiving one global reward. \\
GRPO group & Eight sampled trajectories per prompt. & Rewards are normalized within the prompt group. \\
Reward timing and signals & Computed after the complete rollout from diagnosis and trajectory feedback, plus behavioral constraints. & One reward is assigned to each complete trajectory. \\
Advantage / credit scope & Group-normalized sequence reward broadcast across policy-generated response tokens. & Credit assignment operates at the trajectory level. \\
Optimizer & Standard GRPO implemented with SLIME. & Process awareness is encoded in the reward signals. \\
Train--evaluation separation & Disjoint cases; reward-side terminal generation, extraction, and judging use DS-3.2; held-out reader and evaluator use DS-V3. & Evaluator versions are stage-specific: DS-3.2 for training rewards and DS-V3 for held-out evaluation. \\
\bottomrule
\end{tabular}
\caption{Interactive multi-turn rollout and GRPO update contract. Multi-turn structure enters through environment interaction and trajectory-level reward computation; grouped normalization and policy optimization use standard GRPO.}
\label{tab:multi-turn-grpo-contract}
\end{table*}

Each doctor output is either a question or a structured diagnosis action. The grounded patient simulator responds from case information, and rollouts stop at the diagnosis action or the seven-turn limit. GRPO normalizes grouped rollout rewards and broadcasts the normalized sequence reward across response tokens. The active reward is
\begin{equation}
\begin{aligned}
R ={}& R_{\mathrm{diag}} + R_{\mathrm{traj}}
+ \lambda_{\mathrm{turn}}R_{\mathrm{turn}}
+ \lambda_{\mathrm{format}}R_{\mathrm{format}} \\
&+ \lambda_{\mathrm{rep}}R_{\mathrm{rep}}
+ R_{\mathrm{qcount}}
+ \lambda_{\mathrm{prefix}}R_{\mathrm{prefix}}.
\end{aligned}
\end{equation}
This separate training-time reward targets the diagnosis and trajectory dimensions but is distinct from the held-out $0.5D+0.4T+0.1E$ aggregate Total formula. It also includes behavioral constraints that are absent from the held-out score. Reward ablations remove only diagnosis feedback or trajectory feedback while preserving the simulator, auxiliary constraints, frozen shared diagnostic reader, held-out evaluator, and score coefficients.

\providecommand{\supplementfigurewidth}{\columnwidth}
\begin{figure}[!htbp]
\centering
\includegraphics[width=\supplementfigurewidth]{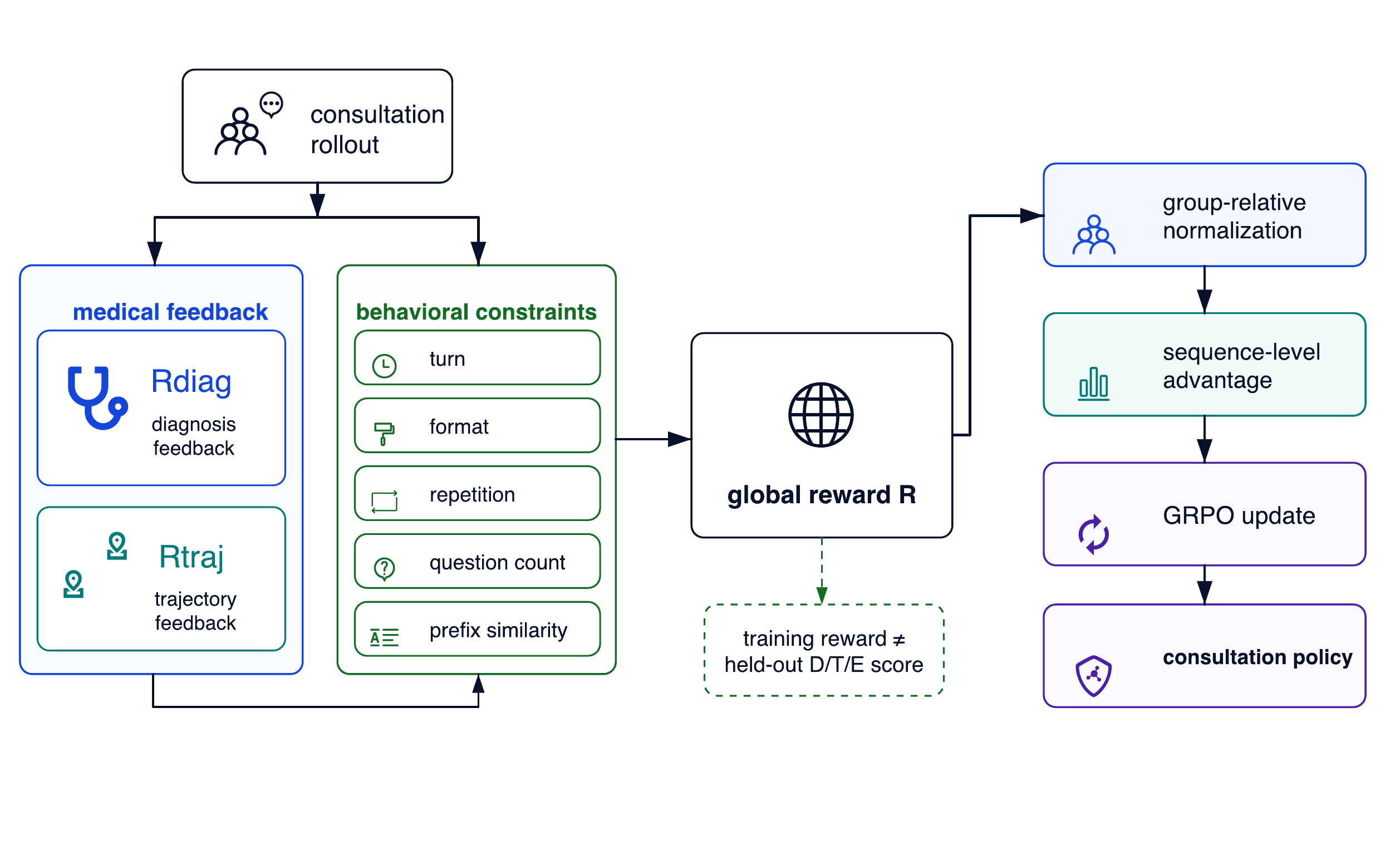}
\caption{Process-aware reward and standard GRPO update. Diagnosis feedback, trajectory feedback, and behavioral constraints form the training reward. Group-relative normalization yields a sequence-level advantage for the policy update; this reward is distinct from the held-out D/T/E score.}
\label{fig:process-aware-reward}
\end{figure}

\begin{table*}[t]
\centering
\small
\setlength{\tabcolsep}{4pt}
\renewcommand{\arraystretch}{1.08}
\resizebox{\textwidth}{!}{%
\begin{tabular}{llll}
\toprule
Component & Implementation variable & Scale / coefficient & Computation summary \\
\midrule
Diagnosis result & \texttt{accuracy\_score} & no outer multiplier & Frozen diagnostic-tool output is compared with the reference diagnosis. \\
Trajectory match & \texttt{reward\_trajectory\_f1} & no outer multiplier & Weighted trajectory F1 from semantic matching of doctor questions to reference targets. \\
Turn budget & \texttt{reward\_turn} & $\lambda_{\mathrm{turn}}=0.8$ & Reward for stopping within the prompt-specific target turn bucket. \\
Format compliance & \texttt{reward\_from\_format} & $\lambda_{\mathrm{format}}=4.0$ & XML/tool-call format score averaged across assistant turns. \\
Repetition control & \texttt{repetition\_score} & $\lambda_{\mathrm{rep}}=0.5$ & Repetition score clipped at 6 before scaling. \\
Question count & \texttt{question\_num\_score} & no outer multiplier & Per-turn question-count compliance with penalties for under- or over-questioning. \\
Prefix control & \texttt{prefix\_penalty} & $\lambda_{\mathrm{prefix}}=0.5$ & Penalty for overly similar response prefixes across turns. \\
\bottomrule
\end{tabular}
}
\caption{Active reward components in the reported process-aware training configuration. Style and session-level total-length terms are disabled, so they are not part of the active reward formula or reward-component ablations. The training reward is distinct from the held-out diagnosis--trajectory--efficiency total score.}
\label{tab:reward-components}
\end{table*}

\section{Main Result Uncertainty and Robustness}

\begin{table}[!htbp]
\centering
\scriptsize
\setlength{\tabcolsep}{2.5pt}
\begin{tabular}{p{0.25\columnwidth}p{0.31\columnwidth}p{0.35\columnwidth}}
\toprule
System group & Control / status & Evidential role \\
\midrule
Qwen3-32B & Starting checkpoint; same simulator, budget, tools, and metrics & Same-family reference for the controlled training contrast \\
Qwen3-32B + GRPO & Process-aware rewards; same held-out splits and score formula & Headline controlled training result \\
GPT-5.4 / GLM-5 & Prompt-only external systems under the common protocol & Context against strong general systems \\
Doctor-R1 / HuatuoGPT-o1 / UltraMedical / Baichuan-M2 & External medical or clinical-inquiry systems & Tests whether specialization alone yields information-acquisition coverage \\
\bottomrule
\end{tabular}
\caption{Baseline protocol and interpretation. Trained same-family comparisons are separated from external references so closed-model comparisons are not treated as optimizer ablations.}
\label{tab:baseline-protocol}
\end{table}

\begin{table*}[t]
\centering
\small
\begin{tabular}{lrrrr}
\toprule
Agent / policy & Diagnosis F1 (\%) & Trajectory F1 (\%) & Efficiency (\%) & Total (\%) \\
\midrule
Qwen3-32B + GRPO & \textbf{44.6{\scriptsize $\pm$3.0}} & 57.2{\scriptsize $\pm$1.7} & \textbf{82.9{\scriptsize $\pm$1.0}} & \textbf{53.4{\scriptsize $\pm$1.3}} \\
GPT-5.4 & 36.3{\scriptsize $\pm$2.3} & \textbf{58.8{\scriptsize $\pm$0.7}} & 79.7{\scriptsize $\pm$0.7} & 49.7{\scriptsize $\pm$1.3} \\
GLM-5 & 38.0{\scriptsize $\pm$7.7} & 54.4{\scriptsize $\pm$1.6} & 66.4{\scriptsize $\pm$16.6} & 47.4{\scriptsize $\pm$5.4} \\
HuatuoGPT-o1-70B & 37.7{\scriptsize $\pm$1.5} & 46.2{\scriptsize $\pm$2.0} & 77.9{\scriptsize $\pm$0.5} & 45.1{\scriptsize $\pm$1.1} \\
Llama-3-70B-UltraMedical & 31.4{\scriptsize $\pm$2.4} & 50.5{\scriptsize $\pm$1.7} & 81.2{\scriptsize $\pm$1.0} & 44.0{\scriptsize $\pm$1.0} \\
Qwen3-32B & 33.1{\scriptsize $\pm$0.9} & 48.1{\scriptsize $\pm$1.3} & 80.2{\scriptsize $\pm$1.1} & 43.8{\scriptsize $\pm$1.0} \\
Baichuan-M2-32B & 32.3{\scriptsize $\pm$3.2} & 46.3{\scriptsize $\pm$2.9} & 81.8{\scriptsize $\pm$1.2} & 42.9{\scriptsize $\pm$2.1} \\
Doctor-R1 & 31.5{\scriptsize $\pm$1.8} & 40.3{\scriptsize $\pm$2.7} & 75.0{\scriptsize $\pm$2.1} & 39.3{\scriptsize $\pm$1.5} \\
\bottomrule
\end{tabular}
\caption{Component results on 100 record-driven held-out cases, ordered by total score. Values are mean $\pm$ sample standard deviation over five repeated runs. Diagnosis and trajectory F1 use strict one-to-one semantic assignment within each run; efficiency uses the first summary round and a lower bound of 0.1 when no valid trajectory item is matched.}
\label{tab:record-results}
\end{table*}

\begin{table*}[t]
\centering
\small
\begin{tabular}{lrrrr}
\toprule
Agent / policy & Diagnosis F1 (\%) & Trajectory F1 (\%) & Efficiency (\%) & Total (\%) \\
\midrule
Qwen3-32B + GRPO & \textbf{41.8{\scriptsize $\pm$2.1}} & \textbf{52.6{\scriptsize $\pm$1.1}} & \textbf{82.4{\scriptsize $\pm$1.2}} & \textbf{50.2{\scriptsize $\pm$1.3}} \\
GPT-5.4 & 35.5{\scriptsize $\pm$2.2} & 52.4{\scriptsize $\pm$0.7} & 79.1{\scriptsize $\pm$0.3} & 46.6{\scriptsize $\pm$1.0} \\
Llama-3-70B-UltraMedical & 39.5{\scriptsize $\pm$3.6} & 46.9{\scriptsize $\pm$2.2} & 79.4{\scriptsize $\pm$1.5} & 46.4{\scriptsize $\pm$2.5} \\
Qwen3-32B & 39.8{\scriptsize $\pm$1.4} & 44.7{\scriptsize $\pm$1.9} & 78.0{\scriptsize $\pm$2.0} & 45.6{\scriptsize $\pm$1.5} \\
Doctor-R1 & 37.4{\scriptsize $\pm$2.4} & 41.2{\scriptsize $\pm$2.6} & 78.6{\scriptsize $\pm$3.4} & 43.1{\scriptsize $\pm$1.5} \\
GLM-5 & 34.3{\scriptsize $\pm$4.2} & 44.7{\scriptsize $\pm$1.7} & 74.9{\scriptsize $\pm$2.0} & 42.5{\scriptsize $\pm$2.2} \\
Baichuan-M2-32B & 33.5{\scriptsize $\pm$2.8} & 44.2{\scriptsize $\pm$1.2} & 80.8{\scriptsize $\pm$2.2} & 42.5{\scriptsize $\pm$1.3} \\
HuatuoGPT-o1-70B & 37.2{\scriptsize $\pm$2.8} & 40.8{\scriptsize $\pm$2.6} & 75.9{\scriptsize $\pm$2.3} & 42.5{\scriptsize $\pm$2.2} \\
\bottomrule
\end{tabular}
\caption{Component results on 70 dialogue-driven held-out cases, ordered by total score. Values are mean $\pm$ sample standard deviation over five repeated runs. Diagnosis and trajectory F1 use strict one-to-one semantic assignment within each run; efficiency uses the first summary round and a lower bound of 0.1 when no valid trajectory item is matched.}
\label{tab:dialogue-results}
\end{table*}

\subsection{Case-Cluster Uncertainty and Weight Sensitivity}

The case-cluster bootstrap resamples the held-out cases available for each comparison with replacement and applies the same draw to every policy and repeated run. Within each draw, it re-aggregates TP/pred/gold to recompute run-level micro D/T, case-mean E, and Total before averaging the five repeats. The resulting point estimates match the main-table contrasts. The 10,000 resamples quantify held-out case-sampling uncertainty for fixed trained checkpoints; training-seed variability lies outside this analysis.

\begin{table*}[t]
\centering
\small
\setlength{\tabcolsep}{4pt}
\begin{tabular}{llrrrrr}
\toprule
Split & Baseline & Cases & Mean diff. & 95\% CI & $P(\Delta>0)$ & W/T/L \\
\midrule
Dialogue & HuatuoGPT-o1-70B & 70 & 7.7 & [5.2, 10.0] & >0.9999 & 47/7/16 \\
Dialogue & Baichuan-M2-32B & 70 & 7.7 & [5.2, 10.2] & >0.9999 & 58/2/10 \\
Dialogue & GLM-5 & 70 & 7.6 & [5.3, 10.0] & >0.9999 & 57/4/9 \\
Dialogue & Doctor-R1 & 70 & 7.1 & [4.8, 9.4] & >0.9999 & 53/2/15 \\
Dialogue & Qwen3-32B & 70 & 4.6 & [2.4, 6.8] & >0.9999 & 50/5/15 \\
Dialogue & Llama-3-70B-UltraMedical & 70 & 3.7 & [1.4, 6.1] & 0.9993 & 43/4/23 \\
Dialogue & GPT-5.4 & 70 & 3.6 & [1.0, 6.2] & 0.9973 & 41/4/25 \\
Record & Doctor-R1 & 100 & 14.1 & [11.7, 16.6] & >0.9999 & 88/4/8 \\
Record & Baichuan-M2-32B & 100 & 10.6 & [8.1, 13.1] & >0.9999 & 72/8/20 \\
Record & Qwen3-32B & 100 & 9.6 & [7.6, 11.8] & >0.9999 & 76/7/17 \\
Record & Llama-3-70B-UltraMedical & 100 & 9.4 & [7.0, 11.8] & >0.9999 & 77/8/15 \\
Record & HuatuoGPT-o1-70B & 100 & 8.3 & [5.9, 11.0] & >0.9999 & 65/8/27 \\
Record & GLM-5 & 100 & 6.0 & [3.7, 8.5] & >0.9999 & 77/4/19 \\
Record & GPT-5.4 & 100 & 3.8 & [1.4, 6.3] & 0.9995 & 54/7/39 \\
\bottomrule
\end{tabular}
\caption{Case-cluster bootstrap over the cases available for each trained-policy comparison. Each resample draws held-out cases and recomputes run-level micro D/T, case-mean E, and Total before averaging the five repeats, so point differences match the main table. Differences and confidence intervals are percentage points from 10,000 resamples. W/T/L is a descriptive count from available repeated-run case means with a one-point tie tolerance.}
\label{tab:current-external-paired-bootstrap}
\end{table*}

\begin{table*}[t]
\centering
\small
\begin{tabular}{lllrrr}
\toprule
Split & Baseline & Component & Mean diff. & 95\% CI & $P(\Delta>0)$ \\
\midrule
Dialogue & GPT-5.4 & Diagnosis & 6.3 & [1.6, 11.1] & 0.9957 \\
Dialogue & GPT-5.4 & Efficiency & 3.3 & [2.1, 4.4] & >0.9999 \\
Dialogue & GPT-5.4 & Trajectory & 0.2 & [-3.0, 3.3] & 0.5390 \\
Dialogue & Qwen3-32B & Diagnosis & 2.0 & [-1.9, 5.8] & 0.8457 \\
Dialogue & Qwen3-32B & Efficiency & 4.4 & [2.1, 6.9] & >0.9999 \\
Dialogue & Qwen3-32B & Trajectory & 7.9 & [5.0, 10.8] & >0.9999 \\
Record & GPT-5.4 & Diagnosis & 8.2 & [4.0, 12.6] & 0.9999 \\
Record & GPT-5.4 & Efficiency & 3.2 & [1.8, 4.6] & >0.9999 \\
Record & GPT-5.4 & Trajectory & -1.7 & [-4.5, 1.2] & 0.1292 \\
Record & Qwen3-32B & Diagnosis & 11.5 & [8.2, 15.0] & >0.9999 \\
Record & Qwen3-32B & Efficiency & 2.7 & [1.0, 4.3] & 0.9990 \\
Record & Qwen3-32B & Trajectory & 9.1 & [6.5, 11.8] & >0.9999 \\
\bottomrule
\end{tabular}
\caption{Case-cluster bootstrap over the cases available for the controlled Qwen3-32B comparison and strongest external total-score reference. Each draw recomputes the run-level component estimators before averaging repeats. Differences are the trained Qwen3-32B + GRPO policy minus baseline in percentage points.}
\label{tab:current-external-component-paired-bootstrap}
\end{table*}

\begin{table*}[t]
\centering
\small
\begin{tabular}{lrrrr}
\toprule
Weights $(D/T/E)$ & Record top & Margin & Dialogue top & Margin \\
\midrule
0.50 / 0.40 / 0.10 & Qwen3-32B + GRPO & 3.76 & Qwen3-32B + GRPO & 3.56 \\
0.55 / 0.35 / 0.10 & Qwen3-32B + GRPO & 4.26 & Qwen3-32B + GRPO & 3.58 \\
0.45 / 0.45 / 0.10 & Qwen3-32B + GRPO & 3.27 & Qwen3-32B + GRPO & 3.25 \\
0.45 / 0.40 / 0.15 & Qwen3-32B + GRPO & 3.51 & Qwen3-32B + GRPO & 3.41 \\
0.50 / 0.35 / 0.15 & Qwen3-32B + GRPO & 4.01 & Qwen3-32B + GRPO & 3.61 \\
0.55 / 0.40 / 0.05 & Qwen3-32B + GRPO & 4.02 & Qwen3-32B + GRPO & 3.71 \\
0.50 / 0.45 / 0.05 & Qwen3-32B + GRPO & 3.52 & Qwen3-32B + GRPO & 3.41 \\
\bottomrule
\end{tabular}
\caption{Local score-weight sensitivity. Margins are percentage-point differences between the top and second system. The trained Qwen3-32B + GRPO policy remains first for all tested weight vectors; coefficients sum to one.}
\label{tab:score-weight-sensitivity}
\end{table*}

\subsection{Alternative Readers}

Table~\ref{tab:diagnostic-tool-robustness} replaces DS-V3 with Qwen3-32B or GPT-5.4 on the same stored consultation histories. Diagnosis is recomputed under the same one-to-one assignment; trajectory and efficiency remain fixed. The trained policy has the highest mean total for both tested readers and both sources. The closest setting supports ordering stability, with a 0.44-point margin that leaves pairwise separation unresolved.

\begin{table*}[t]
\centering
\small
\begin{tabular}{lrrrr}
\toprule
\multirow{2}{*}{Policy} & \multicolumn{2}{c}{Record total (\%)} & \multicolumn{2}{c}{Dialogue total (\%)} \\
\cmidrule(lr){2-3}\cmidrule(lr){4-5}
 & Qwen3-32B reader & GPT-5.4 reader & Qwen3-32B reader & GPT-5.4 reader \\
\midrule
Qwen3-32B + GRPO & \textbf{49.9{\scriptsize $\pm$2.1}} & \textbf{50.4{\scriptsize $\pm$0.5}} & \textbf{46.0{\scriptsize $\pm$0.9}} & \textbf{44.8{\scriptsize $\pm$1.2}} \\
Qwen3-32B & 40.5{\scriptsize $\pm$1.3} & 42.8{\scriptsize $\pm$0.8} & 40.5{\scriptsize $\pm$1.0} & 40.0{\scriptsize $\pm$1.5} \\
GPT-5.4 & 47.4{\scriptsize $\pm$1.8} & 48.1{\scriptsize $\pm$1.2} & 44.1{\scriptsize $\pm$2.3} & 44.3{\scriptsize $\pm$0.7} \\
GLM-5 & 44.5{\scriptsize $\pm$4.3} & 45.6{\scriptsize $\pm$3.7} & 40.0{\scriptsize $\pm$1.2} & 40.6{\scriptsize $\pm$1.4} \\
HuatuoGPT-o1-70B & 41.9{\scriptsize $\pm$0.7} & 42.3{\scriptsize $\pm$1.0} & 38.7{\scriptsize $\pm$3.4} & 39.3{\scriptsize $\pm$2.4} \\
Llama-3-70B-UltraMedical & 43.9{\scriptsize $\pm$0.7} & 45.1{\scriptsize $\pm$1.0} & 41.1{\scriptsize $\pm$1.4} & 42.8{\scriptsize $\pm$1.3} \\
Baichuan-M2-32B & 38.3{\scriptsize $\pm$2.1} & 38.8{\scriptsize $\pm$1.8} & 37.6{\scriptsize $\pm$0.8} & 39.1{\scriptsize $\pm$0.4} \\
Doctor-R1 & 37.8{\scriptsize $\pm$1.5} & 37.2{\scriptsize $\pm$1.0} & 37.7{\scriptsize $\pm$2.4} & 40.3{\scriptsize $\pm$2.4} \\
\bottomrule
\end{tabular}
\caption{Alternative shared diagnostic reader analysis on the same consultation histories. Values are percentage mean $\pm$ sample standard deviation over five runs. Only the frozen post-consultation diagnostic reader changes; trajectory, efficiency, and score weights remain fixed. The trained Qwen3-32B + GRPO policy has the highest total under both readers on both sources.}
\label{tab:diagnostic-tool-robustness}
\end{table*}

\section{Diagnosis--Trajectory Complementarity}

System means show a positive diagnosis--trajectory association, while case--model associations are weak. Split-specific 33rd/67th percentile buckets expose both off-diagonal directions. This supports measurements that are aligned at the model-objective level but non-interchangeable at the individual-output level.

\begin{table*}[t]
\centering
\small
\begin{tabular}{llrrrr}
\toprule
Level & Split & $n$ & Pearson $r$ & Spearman $\rho$ & Off-diagonal HT/LD; LT/HD \\
\midrule
Model means & Record & 8 & 0.566 & 0.238 & -- \\
Model means & Dialogue & 8 & 0.303 & 0.381 & -- \\
Case--model means & Record & 800 & 0.147 & 0.139 & 71; 76 \\
Case--model means & Dialogue & 560 & 0.101 & 0.086 & 50; 63 \\
\bottomrule
\end{tabular}
\caption{Diagnosis--trajectory association under strict one-to-one scoring. System means are positively associated, while case--model associations are weak and both off-diagonal directions remain common under split-specific 33rd/67th percentile thresholds. This supports treating diagnosis and trajectory as related but not interchangeable.}
\label{tab:trajectory-diagnosis-evidence-summary}
\end{table*}

\begin{table*}[t]
\centering
\footnotesize
\setlength{\tabcolsep}{2.5pt}
\begin{tabular}{llrrrrrrr}
\toprule
Split & System & Runs & Cases & Diag. & Traj. & Total & HT/LD $\geq3$ & LT/HD $\geq3$ \\
\midrule
Dialogue & Qwen3-32B + GRPO & 5 & 70 & 44.2 & 53.3 & 51.7 & 4 & 3 \\
Dialogue & Llama-3-70B-UltraMedical & 5 & 70 & 42.6 & 47.9 & 48.4 & 1 & 7 \\
Dialogue & GPT-5.4 & 5 & 70 & 38.6 & 53.0 & 48.4 & 3 & 8 \\
Dialogue & Qwen3-32B & 5 & 70 & 41.8 & 44.5 & 46.5 & 3 & 6 \\
Dialogue & HuatuoGPT-o1-70B & 5 & 70 & 41.2 & 41.7 & 44.8 & 7 & 5 \\
Dialogue & Doctor-R1 & 5 & 70 & 40.5 & 41.5 & 44.7 & 9 & 3 \\
Dialogue & GLM-5 & 5 & 70 & 35.9 & 44.6 & 43.3 & 7 & 5 \\
Dialogue & Baichuan-M2-32B & 5 & 70 & 29.1 & 43.7 & 40.1 & 4 & 4 \\
\midrule
Record & Qwen3-32B + GRPO & 5 & 100 & 45.9 & 58.7 & 54.7 & 6 & 13 \\
Record & GPT-5.4 & 5 & 100 & 38.6 & 59.1 & 50.9 & 7 & 9 \\
Record & HuatuoGPT-o1-70B & 5 & 100 & 39.8 & 46.5 & 46.3 & 6 & 6 \\
Record & Qwen3-32B & 5 & 100 & 33.7 & 48.2 & 44.2 & 4 & 4 \\
Record & Llama-3-70B-UltraMedical & 5 & 100 & 30.0 & 51.4 & 43.7 & 11 & 4 \\
Record & GLM-5 & 5 & 100 & 37.3 & 45.5 & 43.5 & 3 & 2 \\
Record & Baichuan-M2-32B & 5 & 100 & 31.3 & 47.4 & 42.8 & 8 & 6 \\
Record & Doctor-R1 & 5 & 100 & 32.0 & 40.1 & 39.5 & 5 & 4 \\
\bottomrule
\end{tabular}
\caption{Equal-weighted case-level decomposition. Scores are percentages after averaging repeated runs per case. HT/LD and LT/HD count cases assigned to the corresponding model-specific tertile bucket in at least three of five runs. This case-unit view complements, rather than reproduces, the run-level micro leaderboard.}
\label{tab:current-external-case-decomposition}
\end{table*}

\begin{table*}[t]
\centering
\small
\setlength{\tabcolsep}{4pt}
\begin{tabular}{p{2.2cm}p{2.2cm}p{3.2cm}p{4.3cm}p{2.5cm}}
\toprule
Pattern & Bucket frequency & Case-derived setting & Evidence summary & Five-run change \\
\midrule
Stable process improvement & 62/170 (36.5\%) & Acute abdominal pain & The trained policy more consistently covers pain course, associated gastrointestinal symptoms, triggers/history, prior tests, and instability signs; the starting policy is more variable and sometimes repeats narrow symptom checks. & Trajectory $+22.5$ pp; diagnosis $-0.7$ pp \\
\midrule
Diagnosis gain with process shortfall & 7/170 (4.1\%) & Acute jaundice-like presentation & The relevant diagnostic family is recovered more often, while specific exposure, associated-symptom, and history targets remain unelicited; the trained policy's trajectory F1 remains 38.6. & Diagnosis $+43.3$ pp; trajectory $-1.2$ pp \\
\bottomrule
\end{tabular}
\caption{Case-level complementarity under the current protocol-admissible coverage endpoint. The first bucket requires a trajectory gain of at least 10 points with cross-run superiority at least 0.70 and no diagnosis loss above 5 points. The second requires low-trajectory/high-diagnosis behavior in at least three of five trained-policy runs, a diagnosis gain of at least 10 points, and diagnosis superiority at least 0.70. Displayed cases are candidates nearest the multivariate bucket center, not maximum-gain cases. Case descriptions are derived from controlled artifacts and conceptually paraphrased; no raw patient text or identifiers are shown.}
\label{tab:current-case-study-examples}
\end{table*}

\section{Interpretive and Deployment Boundaries}

The semantic judge and frozen shared diagnostic reader are model components of the measurement pipeline. Candidate semantic matches follow protocol-defined directional coverage relations, and D/T/E scores are interpreted under the frozen candidate-generation and matching protocol. Public prompts, matching code, and aggregate results make this dependency inspectable. Alternative shared diagnostic readers characterize dependence on the history-to-diagnosis model, while exhaustive candidate-generation coverage remains a separate measurement question.\ifdefined\ARXIVPREPRINT\else\ The selected-match audit establishes sampled assignment validity, and the blinded D/T/E audit evaluates physician alignment of within-case component ordering.\fi

Trajectory targets are reviewed normative information needs expressed as clinically natural question units. Related subquestions may remain bundled because physicians commonly ask them together; a prediction can receive coverage for one valid component under the prespecified inquiry-type constraints. This design provides coarser within-bundle resolution than atomic-fact recall while allowing semantically compatible questions and removing facts already volunteered by the simulated patient. The targets cover reviewed information needs across flexible consultation strategies. Diagnostic support, information-acquisition coverage, and efficiency should therefore be read together, with efficiency measuring how early matched trajectory targets are covered.

\ifdefined\ARXIVPREPRINT
These boundaries mean that D/T/E should be read as protocol-conditioned research metrics rather than clinical endpoints. Prospective clinical validity, broader reader ensembles, and deployment safety require study designs beyond this offline evaluation.
\else
The two physician audits address different targets. The selected-match audit measures sampled validity of semantic matches awarded credit, and the D/T/E audit measures physician alignment of within-case component ordering. Their protocols, sample sizes, and confidence intervals are reported in the meta-evaluation appendix and Tables~\ref{tab:physician-selected-edge-audit}--\ref{tab:physician-eval-protocol}. These boundaries mean that D/T/E should be read as protocol-conditioned research metrics rather than clinical endpoints. Prospective clinical validity, broader reader ensembles, and deployment safety require study designs beyond this offline evaluation.
\fi

\ifdefined\ARXIVPREPRINT
\else
\input{appendix/b_release}
\fi

\end{document}